\documentclass[10pt,twocolumn,letterpaper]{article}

\usepackage[final]{iccv}      

\definecolor{iccvblue}{rgb}{0.21,0.49,0.74}
\usepackage[pagebackref,breaklinks,colorlinks,allcolors=iccvblue]{hyperref}
\usepackage{todonotes}
\usepackage{multirow}

\usepackage{amssymb}
\usepackage{bbding}

\usepackage{verbatim}

\usepackage[pagebackref,breaklinks,colorlinks,allcolors=iccvblue]{hyperref}
\usepackage{booktabs}

\usepackage{multirow}

\usepackage{amssymb}
\usepackage{bbding}

\usepackage{verbatim}
\usepackage{makecell}
\usepackage{array}

\newcommand{\name}{MVP-LM}
\newcommand{\dname}{Advancing Visual Large Language Model for Multi-granular Versatile Perception}

\title{\dname}

\def\paperID{13968} 
\def\confName{ICCV}
\def\confYear{2025}

\author{Wentao Xiang$^{1}$$^{*}$, Haoxian Tan$^{2}$$^{*}$, Yujie Zhong$^{2}$$^\dagger$, Cong Wei$^{1}$, Dengjie Li$^{2}$, and Yujiu Yang$^{1}$$^{\dagger}$  \\
\textbf{$^{1}$} Tsinghua Shenzhen International Graduate School, Tsinghua University \\
\textbf{$^{2}$} Meituan Inc.\\
{\tt\small thu\_xiangwentao@163.com,} 
{\tt\small jaszhong@hotmail.com,} 
{\tt\small yang.yujiu@sz.tsinghua.edu.cn}
}

\newcommand{\checkQuestions}[1]{\textcolor{black}{#1}}

\begin{document}
\maketitle

\let\thefootnote\relax\footnotetext{
\noindent $^*$ Equal contribution. \quad $^\dagger$ Corresponding author.}

\begin{abstract}
Perception is a fundamental task in the field of computer vision, encompassing a diverse set of subtasks that can be systematically categorized into four distinct groups based on two dimensions: prediction type and instruction type. Notably, existing researches often focus solely on a limited subset of these potential combinations, which constrains their applicability and versatility across various contexts. In response to this challenge, we present \name, a \underline{\textbf{M}}ulti-granular and \underline{\textbf{V}}ersatile \underline{\textbf{P}}erception framework incorporating Visual Large \underline{\textbf{L}}anguage \underline{\textbf{M}}odel. Our framework is designed to integrate both word-based and sentence-based perception tasks alongside box and mask predictions within a single architecture. \name~features an innovative multi-granularity decoder in conjunction with a CoT-inspired dataset unification strategy, enabling seamless supervised fine-tuning across a wide spectrum of tasks, including but not limited to panoptic segmentation, detection, grounding, and referring expression segmentation. Furthermore, we introduce a query enhancement strategy aimed at harnessing the decoding and generative capabilities inherent in VLLMs. Extensive experiments conducted across a range of benchmarks in both word-based and sentence-based perception tasks substantiate the efficacy of our framework. The code will be available \href{https://github.com/xiangwentao666/MVP-LM}{here}.

\end{abstract}    
\section{Introduction}
\label{sec:intro}

Perception constitutes a fundamental task within the field of computer vision, necessitating that models accurately identify and locate objects within images or videos in accordance with specified instructions. These perception tasks can be categorized along two dimensions: prediction type (box vs. mask) and instruction type (word-based vs. sentence-based). This classification yields four different groups.

\begin{table}[t]
  \centering
    \setlength{\belowcaptionskip}{-5pt} 
    \setlength{\abovecaptionskip}{5pt} 
  \caption{The comparison of capabilities of different methods. Current works can only address a subset of the combinations, while our \name~ can cover all tasks.}
  \scalebox{0.65}{
    \begin{tabular}{p{7cm}cccc}
    \toprule[1.1pt] 
     \multirow{2}{*}{Method} & \multicolumn{2}{c}{Output Type} & \multicolumn{2}{c}{Instruction Type} \\
     & Box & Mask & Word & Sentence \\
    \midrule[0.9pt]
    
    \textbf{\emph{Specialists}} \\
    \midrule[0.5pt]
    { OWL\cite{minderer2023owl},  PromptDet\cite{PromptDet_Towards_Open_Vocabulary_Detection_Using_Uncurated_Images}, OV-DINO\cite{wang2024ov_ovdino}} & \Checkmark &  & \Checkmark & \\
    
    X-Decoder\cite{zou2023generalized_xdecoder}, OpenSeeD\cite{zhang2023simple}, Mask DINO\cite{li2023mask} & \Checkmark & \Checkmark & \Checkmark & \multirow{2}{*}{} \\
    
    MDETR\cite{kamath2021mdetr}, GLIP\cite{li2022grounded_glip}, Grounding DINO\cite{liu2024grounding_gdino_groundingdino} & \Checkmark & & \Checkmark & \Checkmark \\
    
    VLT\cite{ding2021vision}, LAVT\cite{yang2022lavt}, PolyFormer\cite{liu2023polyformer} & & \Checkmark & & \Checkmark \\
    \midrule[0.9pt]

    \textbf{\emph{VLLM-based Generalists}} \\
    \midrule[0.5pt] 
    QwenVL\cite{bai2023qwen}, InternVL\cite{chen2024internvl}, DeepseekVL\cite{lu2024deepseek} & \Checkmark &  &  & \Checkmark \\
    LISA\cite{Lai2023LISARS}, PixelLM\cite{Ren2023PixelLMPR}, Glamm\cite{rasheed2023glamm} &  & \Checkmark &  & \Checkmark \\
    OMG-LLaVA\cite{zhang2024omg}, PSALM\cite{zhang2025psalm_psalm}, HyperSeg\cite{hyperSeg} & & \Checkmark & \Checkmark & \Checkmark \\
    \textbf{\name~(Ours)} & \Checkmark & \Checkmark & \Checkmark & \Checkmark \\
 
    \bottomrule[1.1pt]
    \end{tabular}
  }
  \vspace{-5mm}
  \label{tab:task-compare}
\end{table}

The distinction between coordinate-level prediction and pixel-level prediction has been extensively explored in the literature. Given that mask annotations can be converted to box annotations without loss of information, the concurrent optimization of box predictions and mask predictions within the same dataset has emerged as a prominent training paradigm. This approach has demonstrated mutual benefits for both prediction types~\cite{he2017mask, li2023mask}. In contrast to the costly annotations required for segmentation tasks, box annotations are significantly easier to obtain, resulting in a considerably larger dataset. Consequently, another line of research focuses on enhancing segmentation performance by leveraging additional box-annotated data. Numerous methods~\cite{zhang2023simple, tian2021boxinst,li2022box,lan2023vision,lan2021discobox, chen2019hybrid,he2017mask} have been developed to achieve this, ranging from the utilization of pre-trained models to natively annotate box data, to the design of specialized segmentation loss functions tailored for box annotations.

\textit{Word-based perception} employs individual words to denote the categories of targets, which often results in a degree of vagueness and ambiguity. A category name can correspond to none or multiple objects within an image. 
In contrast, \textit{sentence-based perception} utilizes complete sentences to describe the objects of interest, resulting in greater precision. In most cases, only one object within the image satisfies the specified conditions. A common approach in the literature~\cite{liu2024grounding_gdino_groundingdino, li2022grounded_glip, kamath2021mdetr} involves the deployment of a two-stream framework, wherein visual queries extracted by any vision backbone are utilized to decode object locations and compute similarity with the text features generated by a language encoder to identify the objects. However, many of these methods overlook the underlying semantic concepts inherent in the descriptions, resulting in suboptimal outcomes, particularly for sentence-based perception.

Recently, Large Language Models (LLMs) have demonstrated a remarkable ability to comprehend user queries and generate appropriate responses, positioning them as a viable option for sentence-based perception~\cite{bai2023qwen, chen2024internvl,lu2024deepseek, chen2023shikra, lai2024lisa, rasheed2023glamm, ren2024pixellm}. Additionally, LLMs possess the capability to extract language features through forwarding, which aligns well with word-based perception. Therefore, LLMs reveal substantial potential for unifying both perceptions. 

The representative of sentence-based location tasks, Referring Expression Comprehension (REC), has emerged as the standard benchmark for Visual Large Language Models (VLLMs), as box coordinates can be expressed as a sequence of numbers in text form, rendering them compatible with traditional VLLMs. However, many existing models~\cite{bai2023qwen, chen2024internvl, lu2024deepseek} encounter difficulties in achieving pixel-level predictions. Recent advances~\cite{chen2023shikra, lai2024lisa, rasheed2023glamm, ren2024pixellm} have successfully output masks by incorporating an additional mask decoder. Nevertheless, these approaches have not adequately addressed the challenges associated with word-based perception.

Alternatively, other studies~\cite{zhang2024omg, zhang2025psalm_psalm} have demonstrated comparable performance on standard word-based perception benchmarks by employing the VLLM as a versatile decoder. Nonetheless, these models focus solely on pixel-level prediction and overlook box prediction.

As illustrated in \cref{tab:task-compare}, although joint training has been examined for certain combinations of tasks, \textbf{the collaborative effects of joint training across all four groups have not been thoroughly explored in the existing literature.}

To this end, we introduce a \underline{\textbf{M}}ulti-granular and \underline{\textbf{V}}ersatile \underline{\textbf{P}}erception framework incorporating VL\underline{\textbf{LM}}, termed \name, which is capable of detecting and segmenting targeted objects in images according to various types of user instructions within a single model. By employing a Chain-of-Thought (CoT)-inspired data curation~\cite{wei2022chain}, we transform multiple box- and mask-annotated datasets into a unified supervised fine-tuning (SFT) dataset format, thereby accommodating a diverse set of tasks, including panoptic segmentation, object detection, visual grounding, and referring expression segmentation. Furthermore, to leverage the decoding and generative capabilities of LLMs, we enhance the queries for object identification and localization by incorporating features derived from LLM-generated sequences.

In summary, our contributions are outlined as follows:

\begin{itemize}
    \item We introduce \name, a framework that harnesses both the decoding and generative capabilities of VLLM to perform both \textit{word-based} and \textit{sentence-based} perception tasks across varying granularities.
    \item We have developed a CoT-inspired data curation that unifies datasets from diverse tasks into a single SFT dataset, thereby encouraging the model to adopt a "thinking-then-perceiving" paradigm.
    \item \name~ demonstrates competitive performance across various benchmarks in both word-based and sentence-based perception, thereby validating the potential of our framework. Notably, our \name~achieves remarkable results on both closed-set and open-set tasks compared with other specialists and VLLMs.
\end{itemize}

\begin{figure*}[t]
    \centering
    \includegraphics[width=0.9\textwidth, trim=0in 2.1in 0in 2.2in, clip]{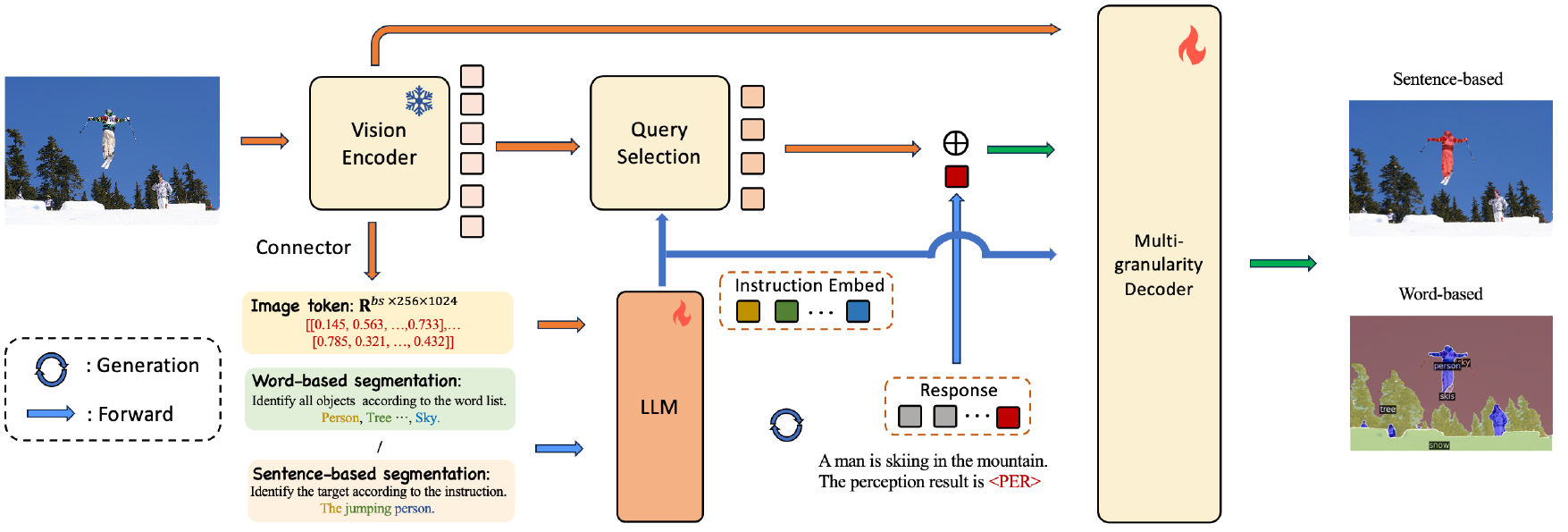}
    \caption{\textbf{Overview of \name}. \name~ implements perception by integrating a multi-granularity decoder into the existing VLLM framework. We utilize a unified prompt template to construct the input sequence for the LLM across different tasks. The base query is derived from the summary token of the generated response. Concurrently, we extract the instruction embeddings from the input sequence (denoted by \textcolor{blue}{the same color}) and select the corresponding residuals from the multi-scale visual features based on these embeddings. After aggregating the base query vector with the residuals, we decode the bounding box and segmentation mask by inputting the resulting queries into the multi-granularity decoder.}
    \label{fig:arch}
    \vspace{-5mm}
\end{figure*}
\section{Related Work}
Perception in computer vision can be categorized into four groups along two dimensions: prediction type and instruction type.
Below, we review both the related work within each of these four groups and CoT, and highlight the gaps \name~aims to fill.

\subsection{Word-based Preception}
\noindent \textbf{Detection and Grounding.} 
Word-based location tasks encompass object detection, grounding, and instance detection, focusing on identifying and localizing objects via bounding boxes with categorical labels. Conventional frameworks~\cite{vild, BARON, dk-detr, fvlm, fu2024frozen-detr, TOOD_Task_aligned_One_stage_Object_Detection, PromptDet_Towards_Open_Vocabulary_Detection_Using_Uncurated_Images, instagen, MrDETR} utilize CLIP-like encoders to compute visual-linguistic similarities, enabling zero-shot capabilities. Significant advances include GLIP~\cite{li2022grounded_glip}, which unifies phrase grounding with object detection, and OV-DINO~\cite{wang2024ov_ovdino}, which implements an end-to-end training pipeline for enhanced detection performance and generalization capability.

\noindent \textbf{Segmentation.} 
Word-based segmentation tasks, encompassing panoptic, semantic, and instance segmentation, demand precise pixel-level predictions. However, generating such masks from textual inputs remains underexplored, largely due to the scarcity of fine-grained annotated data. Existing methods~\cite{zou2023generalized_xdecoder, li2022fully,zhang2023simple,xiong2019upsnet,carion2020end,wang2021max, Open_Vocabulary_Semantic_Segmentation_with_Decoupled_One_Pass_Network, vCLR} primarily align word embeddings with visual features, resulting in coarse segmentation masks. Nevertheless, these approaches~\cite{zou2023generalized_xdecoder,zhang2023simple} often demonstrate limited textual understanding and lack the requisite fine-grained control for high-quality mask generation, which ultimately restricts their performance in both closed-set and open-set tasks.

\subsection{Sentence-based Preception}
\noindent \textbf{Referring Expression Comprehension.}
Sentence-based location tasks, which require a deeper multi-modality understanding, have become a standard benchmark for VLLMs. Specifically, Referring Expression Comprehension (REC) is often the preferred task, as box coordinates can be represented as a text sequence of numbers, making them compatible with conventional VLLMs~\cite{liao2020real, liu2020learning,zhang2018grounding,liu2019knowledge,rohrbach2016grounding,wang2019neighbourhood,yang2019dynamic, bai2023qwen, chen2024internvl, lu2024deepseek}. However, many existing models struggle to achieve pixel-level predictions, and even more so, to facilitate joint training with sentence-based mask perception.

\noindent \textbf{Referring Expression Segmentation.}
Sentence-based segmentation tasks demand a fine-grained understanding of descriptive sentences to accurately generate segmentation masks for target objects. Traditional approaches~\cite{chen2019see,ding2020phraseclick,hu2020bi,huang2025denselyconnectedparameterefficienttuning_DETRIS,nguyentruong2024visionawaretextfeaturesreferring_traditional_referring_image_segmentation_VATEX,huang2020referring,hui2020linguistic} have made strides in fusing visual and textual features to improve segmentation performance, but often fall short in fully leveraging the richness of sentence descriptions, resulting in suboptimal performance in both closed-set and open-set scenarios. 
Recent efforts~\cite{liu2023gres_grefcoco, chen2023shikra, lai2024lisa, rasheed2023glamm, ren2024pixellm} have successfully implemented Referring Expression Segmentation (RES) by incorporating an additional mask decoder. However, these approaches fall short in addressing word-based perception.

\subsection{Chain-of-Thought}
Chain-of-Thought prompting \cite{wei2022chain} was initially introduced during inference to enhance the reasoning capabilities of LLMs. Subsequently, CoT has evolved into a prevalent paradigm in the post-training phase, wherein models first undergo supervised fine-tuning on CoT datasets, followed by reinforcement learning~\cite{shao2024deepseekmathpushinglimitsmathematical}. A similar trend has also been observed in VLLMs, which have demonstrated promising performance in multi-modal understanding and reasoning tasks. Various meticulously designed rewards~\cite{yang2025r1onevisionadvancinggeneralizedmultimodal, liu2025visualrftvisualreinforcementfinetuning, zhang2025r1vllearningreasonmultimodal, 2025arXiv250313377W, R1-Omni, feng2025videor1reinforcingvideoreasoning}, ranging from format to geometry, have been proposed to handle diverse scenarios.

\subsection{Identified Gaps and Our Contribution}
While significant progress has been made in addressing some of the combinations~\cite{chen2024internvl, lu2024deepseek, lai2024lisa, ren2024pixellm, zhang2025psalm_psalm}, a comprehensive solution effectively handling all four groups remains elusive. \name~bridges this gap by offering a solution capable
of simultaneous perception across all four combinations,
marking a significant advancement in the field, addressing the current limitations, and setting a new benchmark for
multi-dimensional visual perception.

\section{Methodology} \label{sec:method}
This section provides an in-depth description of our proposed methodology. We begin by detailing the architecture of our \name~in \cref{subsect: mllm_architecture_design}, followed by an exposition of our strategies for dataset processing in \cref{subsect: dataset_enhancement}.

\subsection{\name~Architecture Design}\label{subsect: mllm_architecture_design}

\textbf{System Overview.} As illustrated in \cref{fig:arch}, our architecture comprises four essential components: an image encoder (Swin-Base Transformer~\cite{liu2021swin}), a connector module, an LLM (Phi-1.5~\cite{gunasekar2023textbooks_phi1_5}), and a multi-granularity decoder. The image encoder extracts rich visual features from input images, serving as the foundation for subsequent processing. The connector module facilitates seamless integration between visual and textual modalities by aligning features from the image encoder with the token embeddings for the LLM. The multi-granularity decoder is responsible for producing both bounding boxes and segmentation masks. 

\begin{figure}[t]
    \centering
    \includegraphics[width=0.48\textwidth, trim=1.5in 2.5in 1in 2.5in, clip]{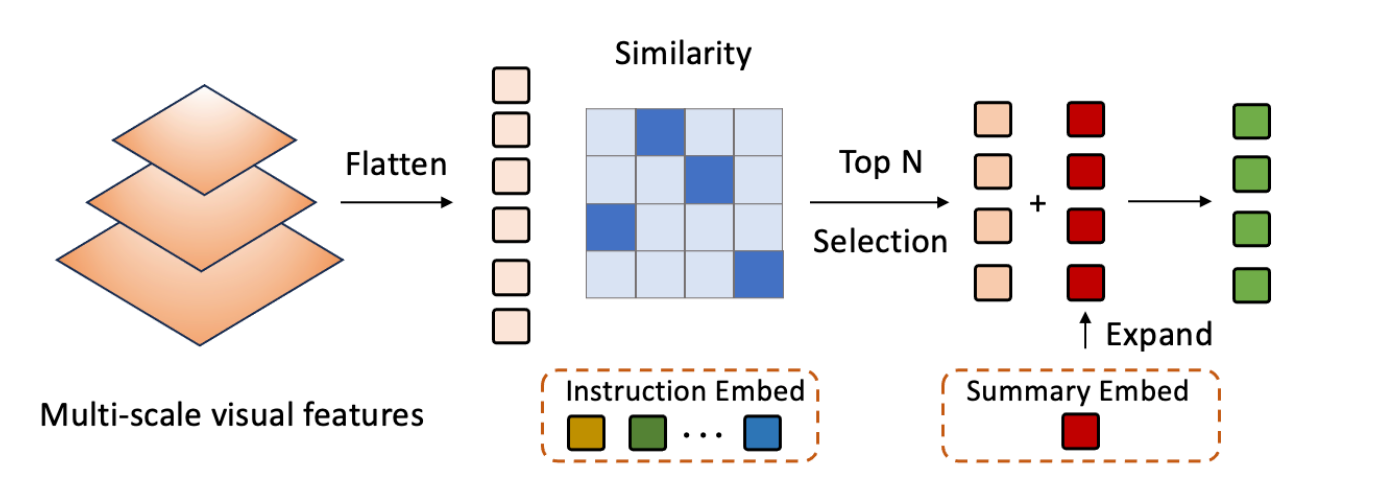}
    \caption{\textbf{Dynamic query selection}. 
    \name~expands the embedding (hidden states) of the summary token into $N$ \textcolor{red}{base query vectors}. Concurrently, multi-scale visual features are flattened, and their similarity with the instruction embedding from the input sequence is computed. The top $N$ \textcolor{orange}{similar features} are then selected and integrated with the corresponding $N$ base query vectors, resulting in $N$ \textcolor{OliveGreen}{final queries} tailored for multi-granularity perception tasks. This dynamic selection mechanism ensures that the most relevant visual features are incorporated into each query, enhancing the precision and adaptability of the decoder.}
    \label{fig:mixed_query}
    \vspace{-5mm}
\end{figure}

\noindent\textbf{Dynamic Query Generation.}
The dynamic generation of visual queries is pivotal for the model's ability to handle diverse perception tasks effectively. As depicted in \cref{fig:mixed_query}, each visual query is composed of two components: a context-aware base query vector and language-guided residuals. 

Initially, \name~leverages the VLLM to generate a specialized summary token that encapsulates the essential information from the input sequences. The hidden states corresponding to this summary token are projected into the base query vectors using an MLP, providing a robust foundation for query formulation.

Simultaneously, we compute the similarity between multi-scale visual features and the text embedding derived from the input instructions (i.e., category names for word-based perception and referring expressions for sentence-based perception). To facilitate this comparison, an MLP aligns the dimensionality of the text embeddings with that of the multi-scale visual features. The top $N$ most similar visual features are then identified, ensuring that the queries are informed by the most relevant visual information. These selected features are integrated with the corresponding base query vectors, resulting in $N$ refined final queries that encapsulate both contextual and visual clues. This dynamic query generation process enables \name~to adaptively focus on relevant regions of the image, thereby enhancing the model's ability to generate accurate bounding boxes and segmentation masks.

\noindent\textbf{Multi-granularity Decoder.}
The multi-granularity decoder is the cornerstone of our architecture, facilitating the simultaneous output of bounding boxes and segmentation masks. Our decoder is inspired by the design principles of OpenSeeD~\cite{zhang2023simple}, which emphasizes the profound integration of multi-scale features and comprehensive query processing. Initially, a set of content queries and their corresponding reference points are generated through a query selection mechanism as seen in \cref{fig:mixed_query}. These queries are then iteratively cross-attended with multi-scale visual features via deformable attention layers, enabling the decoder to capture both global contextual information and detailed local features. The outputs from each MSDeform layer are subsequently processed by three shared heads—a cross-modal similarity computation head, a box head, and a mask head—to produce predictions. The unified output streamlines the training process and allows the model to learn shared representations that benefit both detection and segmentation tasks.

\checkQuestions{}

\noindent\textbf{Training Objectives.} To fully exploit the capabilities of \name, we employ a joint training approach that simultaneously optimizes multiple tasks across diverse datasets. The training objective is formulated as a composite loss function comprising several components:
\begin{align*}
\mathcal{L}= \mathcal{L}_{llm}&+\lambda_{word}\mathcal{L}_{word}+\lambda_{sent}\mathcal{L}_{sent}\\
&+\mathcal{L}_{mask}+\mathcal{L}_{box}
\end{align*}
Here, $\mathcal{L}_{llm}$ represents the standard loss for next-token prediction, ensuring that the LLM accurately generates coherent and contextually appropriate summaries. The term $\mathcal{L}_{word}$ corresponds to the Cross-Entropy (CE) loss for word-based perception conditions, facilitating precise classification based on single-word instructions. Conversely, $\mathcal{L}_{sent}$ denotes the Binary Cross Entropy (BCE) loss for sentence-based perception conditions, enabling the model to handle more complex, descriptive instructions effectively. The mask loss, $\mathcal{L}_{mask}$, is composed of a pixel-level BCE loss and a Dice loss, which together promote accurate and smooth segmentation masks by penalizing discrepancies between predicted and ground-truth masks at both the pixel and region levels. The box loss, $\mathcal{L}_{box}$, includes the $l1$ loss to minimize the absolute differences between predicted and ground-truth bounding box coordinates, as well as the Generalized Intersection over Union (GIoU) loss to enhance the alignment and overlap between predicted and actual boxes. Formally, these loss components are defined as:
\begin{align*} 
\mathcal{L}_{mask}&=\lambda_{bce}\mathcal{L}_{bce}+\lambda_{dice}\mathcal{L}_{dice},\\
\mathcal{L}_{box}&=\lambda_{l1}\mathcal{L}_{l1}+\lambda_{giou}\mathcal{L}_{giou},
\end{align*}

Following the methodology of OpenSeeD, we employ bipartite matching to match the model's predictions with their corresponding ground truth annotations during training. \checkQuestions{Predictions from each layer are assigned to their corresponding annotations to calculate losses through Hungarian matching. Moreover, a denoising strategy is incorporated, similar to that used in MaskDINO, to stabilize optimization and accelerate convergence. However, during inference, only the final layer's output is utilized.}

\subsection{Data Curation}\label{subsect: dataset_enhancement}
\noindent\textbf{CoT-inspired Dataset Unification.} We unified heterogeneous datasets into an SFT dataset specially tailored for the training of VLLM. For each sample in the perception datasets, we construct a question–answer pair based on the original annotations. Each training pair consists of three key components: a task description, an instruction, and a response that embodies a "thinking-then-perceiving" process.

The task description specifies the particular task associated with the data. In the case of word-based perception, the description may be articulated as follows: \textit{"Please identify all objects according to the given phrase list. This is all the candidate phrases."} Conversely, for sentence-based perception, the description is framed as: \textit{"Please identify the target according to the following instruction."}

The instruction varies according to the specific tasks involved. For word-based perception, we concatenate the provided word lists into a single sentence, using a comma as a separator. For instance, given three categories: apple, banana, and orange, the resultant sentence would be \textit{"apple, banana, orange."} After passing this sentence through VLLM, we select the corresponding output embeddings and apply average pooling to obtain the text embedding for each word, while for sentence-based perception, the referring expression directly serves as the final instruction, requiring no additional processing. The text embedding is derived simply by applying pooling over the entire expression.

The response format is standardized across various tasks. We adopt a "thinking-then-perceiving" strategy by prepending the input image caption to the summary, thereby encouraging the VLLM to effectively capture image details. Specifically, the response is structured as follows: \textit{"[image caption]. The perception result is \textless PER\textgreater"}, where \textit{"\textless PER\textgreater"} denotes a special token representing the summary. The hidden states associated with this summary token are subsequently utilized to derive the visual queries.

\noindent\textbf{Multi-caption Auto-labeling.}
To fully leverage the generative prowess of VLLMs, we design our system to generate comprehensive image captions in conjunction with summary tokens. This generation approach enriches the contextual information available to the model, facilitating a deeper understanding of the visual content. However, a significant challenge arises as the majority of our training datasets lack image captions. To address this, we integrate open-source VLLMs to autonomously generate descriptive captions for each image prior to the training phase.

Specifically, we employ a diverse set of VLLMs, including VILA-3B, VILA-13B, InternVL2-8B, and InternVL2-26B, to produce multiple unique captions for the same image. The utilization of distinct VLLMs not only enhances the accuracy of the generated captions but also introduces substantial diversity, mitigating the risk of model overfitting to a single captioning style or vocabulary. By incorporating multi-caption auto-labeling, our approach ensures a more comprehensive and resilient training paradigm, effectively bridging the gap between visual inputs and linguistic descriptions.

\noindent\textbf{Data Refinement.} During the data generation process, the occurrence of irregular or erroneous data is inevitable. Consequently, we implement a straightforward data-cleaning procedure for the dataset. First, we utilize the length and information entropy of the generated captions to filter out meaningless results. Second, we remove sentences containing suggestive keywords such as \textit{"indicate", "might", "may", and "imply"}, as well as other similar terms, to mitigate the hallucinations that are inherent in large language models. These two mechanisms ensure the quality of the generated captions.

\noindent\textbf{Randomness in Training Pairs.} The variability of training pairs is essential for preventing the model from learning spurious correlations and for enhancing its generalization capabilities. In our framework, such diversity is achieved by randomly selecting one of four generated captions for each training response. However, fixed category name ordering in word-based datasets can still introduce spurious causal relationships, leading to overfitting and impaired performance on unseen data. 

To counteract these issues, our mitigation strategy is twofold. Firstly, we introduce randomness by shuffling the order of the word list in each iteration. This disrupts fixed positional biases, preventing the model from relying on word order for predictions. Secondly, word lists are dynamically constructed by randomly incorporating a subset of negative category names (categories absent from the image). This selective inclusion diversifies training prompts, proving particularly beneficial for datasets with extensive vocabularies. By carefully balancing positive and negative categories within VLLM input sequence length constraints, the model's ability to discern relevant from irrelevant objects is enhanced. This approach fosters a more adaptive learning process, significantly reducing overfitting and ensuring robust performance across both closed-set and open-set detection scenarios.

\begin{table}[!t]

    \setlength{\belowcaptionskip}{-5pt} 
    \setlength{\abovecaptionskip}{5pt} 
\caption{Hyperparameters for both training stages. "CP", "RC", "O", and "GG" denote COCO-Panoptic, RefCOCO, Objects365, and GoldG, respectively.}
\label{tab:implementation_details_in_rebuttal}
\centering
\small
\resizebox{0.48\textwidth}{!}{
    \begin{tabular}{>{\centering\arraybackslash}m{0.4\linewidth}
                >{\centering\arraybackslash}m{0.5\linewidth}
                >{\centering\arraybackslash}m{0.5\linewidth}}
        \toprule
        \textbf{Parameters} & \textbf{Stage1} & \textbf{Stage2} \\
        \midrule
        Training Components & Connector & Connector + LLM + Multi-granularity Decoder \\
        Optimizer & \multicolumn{2}{c}{AdamW} \\
        Training Rate & $2 \times 10^{-3}$ & $4 \times 10^{-5}$ \\
        Batch Size & 128 & 64 \\
        Number of Steps & 4650 & 80000 \\
        Learning Rate Schedule & \multicolumn{2}{c}{Cosine Decay} \\
        Weight Decay & 0.0 & 0.05 \\
        Warmup Ratio & \multicolumn{2}{c}{0.03} \\
        Training Data & CC3M & CP(33.3\%) RC(33.3\%) O(16.7\%) GG(16.7\%) \\
        \raisebox{-1.2em}{Loss} & \raisebox{-1.2em}{$L_{LLM}$} & \parbox[t]{\linewidth}{
            \centering
            $L_{LLM} + 2 \cdot L_{word/sent} +$\\[0em]
            $5 \cdot L_{L1} + 2 \cdot L_{GIoU} +$\\[0em]
            $5 \cdot L_{BCE} + 5 \cdot L_{DICE}$
        } \\
        Image Size & \multicolumn{2}{c}{1024 $\times$ 1024} \\
        Image Processing & \multicolumn{2}{c}{\makecell{Resize longer to 1024 and pad shorter to 1024}} \\
        \bottomrule
    \end{tabular}
}
\vspace{-6mm}
\end{table}

\begin{table*}[t]
  \centering
    \setlength{\belowcaptionskip}{-5pt} 
    \setlength{\abovecaptionskip}{5pt} 
    \caption{ Quantitative results on the closed-set COCO-Panoptic segmentation and open-set segmentation tasks. Our model \name~achieves remarkable performance among previous VLLM-based methods.}
\scalebox{0.81}{
  \begin{tabular}{c|l|c|cc|cc|c|c|c}
    \toprule[1.1pt]
    \multirow{2}{*}{Type} & \multirow{2}{*}{~~~~~~~~~Method} & \multirow{2}{*}{Backbone} & \multicolumn{2}{c|}{COCO-Panoptic} & \multicolumn{2}{c|}{ADE-OV} & \multicolumn{1}{c|}{Citys-OV} & \multicolumn{1}{c|}{PC59-OV} & \multicolumn{1}{c}{PAS20-OV} \\
    & & & PQ & mIoU & PQ & mIoU & PQ & mIoU & mIoU  \\
    \midrule[0.7pt]
    \multirow{7}{*}
    {\shortstack{Specialists}}  &  Mask2former~\cite{cheng2022masked} & Swin-B  & 55.1 &  65.1 & - & - & - & - & - \\
    & OneFormer~\cite{jain2023oneformer} & Swin-L  & 57.9  & 67.4 & - & - & - & - & - \\
    & OpenSeeD~\cite{zhang2023simple} & Swin-L  & 59.5  & 68.6 & 19.7 & 23.4 & 41.4 & - & - \\
    & SEEM~\cite{seem}  &  DaViT-B & 56.1  & 66.3 & - & - & - & - & -  \\
    & MaskCLIP~\cite{maskclip} & ViT-L & 30.9 & 47.6 & 15.1 & 23.7 & - & 45.9 & - \\

    & SimBaseline~\cite{xu2022simple} & ViT-B & - & - & - & 20.5 & - & 47.7 & 88.4 \\

    & DeOP~\cite{Open_Vocabulary_Semantic_Segmentation_with_Decoupled_One_Pass_Network} &  ResNet-101c & - & - & - & 22.9 & - & 48.8 & 91.7 \\
    & DaTaSeg~\cite{gu2024dataseg}   &  ViTDet-B  & 52.8  & 62.7 & 12.3 & 18.3 & 28.0 & 51.1 & - \\
    
    \midrule[0.5pt]
    \multirow{3}{*}{\shortstack{VLLM-based\\Generalists}}
    & OMG-LLaVA ~\cite{zhang2024omg} & ConvNeXt-L & 53.8 &  - & - & - & - & - & - \\
    & PSALM ~\cite{Lai2023LISARS} & Swin-B & 55.9 & 66.6 & 13.7 & 18.2 & 28.8 & \textbf{48.5} & 81.3 \\
    & \textbf{\name} & Swin-B & \textbf{56.1} & \textbf{66.8} & \textbf{19.4} & \textbf{20.5} & \textbf{35.3} & 44.1 & \textbf{85.7} \\
    \bottomrule[1.1pt]
  \end{tabular}
}
  \label{tab:cocoseg}
\end{table*}

\begin{table*}[t]
  \centering
    \setlength{\belowcaptionskip}{-5pt} 
    \setlength{\abovecaptionskip}{5pt} 
  \caption{ Comparison with the state-of-the-art models on the closed-set Referring Segmentation benchmarks (RefCOCO series) and more challenging Generalized Referring Expression Segmentation benchmark gRefCOCO\cite{liu2023gres_grefcoco}. \textcolor{gray}{Gray} numbers denote the method using gRefCOCO for training.
  }
  \scalebox{0.84}{
    \begin{tabular}{c|l|ccc|ccc|cc|ccc}
    \toprule[1.1pt] 
    \multirow{2}{*}{ Type } & \multirow{2}{*}{ ~~~~~~~~Method } & \multicolumn{3}{c|}{ RefCOCO } & \multicolumn{3}{c|}{ RefCOCO+ } & \multicolumn{2}{c|}{ RefCOCOg } & \multicolumn{3}{c}{ gRefCOCO }\\
    & & val & testA & testB & val & testA & testB & val(U) & test(U) & val & testA & testB\\
    \midrule[0.7pt]
    \multirow{4}{*}{\shortstack{Specialists}} 
    & VLT~\cite{ding2021vision} & 67.5 & 70.5 & 65.2 & 56.3 & 61.0 & 50.1 & 55.0 & 57.7 & \textcolor{gray}{52.5} & \textcolor{gray}{62.2} & \textcolor{gray}{50.5} \\
    & CRIS~\cite{wang2022cris} & 70.5 & 73.2 & 66.1 & 62.3 & 68.1 & 53.7 & 59.9 & 60.4 & \textcolor{gray}{55.3} & \textcolor{gray}{63.8} & \textcolor{gray}{51.0} \\
    & LAVT~\cite{yang2022lavt} & 72.7 & 75.8 & 68.8 & 62.1 & 68.4 & 55.1 & 61.2 & 62.1 & \textcolor{gray}{57.6} & \textcolor{gray}{65.3} & \textcolor{gray}{55.0} \\
    & PolyFormer-B~\cite{liu2023polyformer} & 74.8 & 76.6 & 71.1 & 67.6 & 72.9 & 59.3 &67.8 &69.1 & - & - & - \\
    \midrule[0.5pt] 
    \multirow{7}{*}{\shortstack{VLLM-based\\Generalists}} & LISA-7B~\cite{Lai2023LISARS} & 74.1 & 76.5 & 71.1 & 62.4 & 67.4 & 56.5 &  66.4 & 68.5 & \textcolor{gray}{38.7} & \textcolor{gray}{52.6} & \textcolor{gray}{44.8} \\
    & PixelLM-7B~\cite{Ren2023PixelLMPR}& 73.0 &76.5 & 68.2 & 66.3 & 71.7 & 58.3 &69.3 & 70.5 & - & - & - \\
    & GSVA-7B~\cite{xia2023gsva}& 76.4 &77.4 &72.8 &64.5 &67.7 & 58.6 & 71.1 & 72.0 & \textcolor{gray}{61.7} & \textcolor{gray}{69.2} & \textcolor{gray}{60.3} \\
    & LaSagnA-7B~\cite{wei2024lasagna} & 76.8 & 78.7 & 73.8 & 66.4 & 70.6 & 60.1 & 70.6 & 71.9 & 38.1 & 50.4 & 42.1 \\
    & OMG-LLaVA ~\cite{zhang2024omg} & 78.0   & 80.3  & 74.1   & 69.1    & 73.1     & 63.0   & 72.9     & 72.9  & - & - & -  \\
    & PSALM ~\cite{zhang2025psalm_psalm} & 83.6   & 84.7  & 81.6   & 72.9    & 75.5     & \textbf{70.1}   & 73.8     & 74.4 & \textbf{42.0 }& \textbf{52.4} & \textbf{50.6}  \\ 
 
    & \textbf{\name} & \textbf{83.6} & \textbf{85.1} & \textbf{82.5} & \textbf{73.9} & \textbf{76.8} & 65.4 & \textbf{75.1} & \textbf{75.6} & 37.6 & 50.4 & 49.9 \\
    \bottomrule[1.1pt]
    \end{tabular}
}
\vspace{-2mm}
\label{tab:ref}
\end{table*}

\section{Experiments}

\textbf{Datasets.} We train \name~in a multi-dataset and multi-task manner like~\cite{gu2024dataseg,zhang2025psalm_psalm}. For generic perception, we use COCO Panoptic Segmentation~\cite{Lin2014MicrosoftCC_coco}(with 118K samples in total) and O365~\cite{shao2019objects365}(with 1.7M samples in total). We use RefCOCO series~\cite{yu2016modeling,nagaraja2016modeling}(with 126K samples in total) for referring segmentation tasks while utilizing the GoldG~\cite{kebe2021a_goldg}(with 770K samples in total) dataset for grounding.
For evaluation, we report the main results on conventional COCO image segmentation tasks (panoptic/semantic) and RefCOCO series. Besides, we conduct evaluation on more challenging perception tasks like open-set segmentation and detection~\cite{zhou2019semantic_ade20k,Cordts2016Cityscapes,mottaghi2014role,everingham2010pascal}.

\noindent \textbf{Evaluation Metrics.}
We report results in widely used evaluation metrics: panoptic quality (PQ) and mean intersection-over-union (mIoU) for panoptic and semantic segmentation, respectively. Additionally, we use cumulative Intersection-over-Union (cIoU) for referring segmentation evaluation consistent with previous studies.

\noindent \textbf{Implementation Details.} 
\checkQuestions{
The model architecture integrates Phi1.5~\cite{gunasekar2023textbooks_phi1_5} as the language model and Swin-B~\cite{liu2021swin} as visual encoder, with our multi-granularity decoder building on OpenSeeD~\cite{zhang2023simple}. For most tasks, we initialize 100 queries for subsequent mask decoding, and for panoptic segmentation, we introduce an additional 100 learnable queries specifically for stuff categories.
Our training proceeds in two phases following LLaVA~\cite{liu2024llava}: 1) visual-language alignment trains exclusively the connector on CC3M~\cite{sharma2018conceptual} while freezes other components; 2) subsequent fine-tuning stage optimizes the whole model except the vision encoder. Detailed hyperparameters and training configurations are listed in \cref{tab:implementation_details_in_rebuttal}. Training spans 80k iterations on the aggregated datasets and 9k for ablation studies, which utilize only the COCO Panoptic and RefCOCO datasets.
}

\subsection{Main results}
\textbf{Word-based Perception Results.} 
We present the performance of \name~on the COCO-Panoptic~\cite{Lin2014MicrosoftCC_coco} and open-set segmentation~\cite{zhou2019semantic_ade20k,Cordts2016Cityscapes,mottaghi2014role,everingham2010pascal} tasks as detailed in \cref{tab:cocoseg}. \name~demonstrates competitive performance when compared to both specialized models and VLLM-based methods in closed-set tasks. Notably, it exhibits exceptional performance in the realm of open-set segmentation within the context of VLLM. Specifically, when evaluated on the open-set ADE20K dataset~\cite{zhou2019semantic_ade20k}, our \name~demonstrates substantial improvements over the VLLM-based PSALM~\cite{zhang2025psalm_psalm} and achieves gains of 5.7 and 2.3 in PQ and mIoU metrics, respectively. Additionally, it surpasses PSALM by 6.5 and 4.4 on the mIoU metric for the open-set datasets Cityscapes~\cite{Cordts2016Cityscapes} and PascalVOC-20~\cite{everingham2010pascal}, respectively.

\noindent \textbf{Sentence-based Perception Results.} \checkQuestions{We evaluate \name~against state-of-the-art methods on the referring expression segmentation (RES) benchmarks, RefCOCO/+/g\cite{yu2016modeling_refcoco_plus_g, nagaraja2016modeling_refcoco_plus_g} as well as the more challenging generalized referring expression segmentation benchmark, gRefCOCO\cite{liu2023gres_grefcoco} in \cref{tab:ref}. We also include RefCOCO series for referring expression comprehension (REC) in \cref{tab:rec_result}. In the domain of RES, \name~achieves state-of-the-art performance on the RefCOCO and RefCOCOg datasets across all subsets. In the RefCOCO+ dataset, \name~attains scores of 73.9 and 76.8 on val and testA, respectively. Moreover, our \name, only 1.3B parameters, outperforms listed 7B and 13B models, and achieves metrics that are second only to DeepSeek-VL2 with over 200B parameters in REC. These results highlight the robustness and generalization capabilities of \name~when compared to previous methods.}

\setlength{\floatsep}{14pt} 

\begin{table}[t]
    \centering
    \setlength{\belowcaptionskip}{-5pt} 
    \setlength{\abovecaptionskip}{5pt} 
    \caption{
       \checkQuestions{ Results on referring expression comprehension tasks. DS-VL2 is the abbreviation for DeepSeek-VL2.} 
    }
    \resizebox{0.48\textwidth}{!}{
  
    \begin{tabular}{c|c|ccc|ccc|cc}
    \toprule[1.1pt] 
    \multirow{2}{*}{ Method } & \multirow{2}{*}{ Params } & \multicolumn{3}{c|}{ RefCOCO } & \multicolumn{3}{c|}{ RefCOCO+ } & \multicolumn{2}{c}{ RefCOCOg } \\
    & & val & testA & testB & val & testA & testB & val & test \\
    \midrule[0.7pt]
    
    Shikra & 13B & 87.8 & 91.1 & 81.8 & 82.9 & 87.8 & 74.4 & 82.6 & 83.1 \\
    MiniGPTv2 & 7B & 88.7 &  91.7 &  85.3 &  80.0 & 85.1  &  74.5 &  84.4 &  84.7 \\
    Octopus & 7B & 89.0 &  92.6 &  83.4 &  83.5 & \textbf{89.4} & 76.0 & 84.4 & 86.2 \\

    \textbf{MVP-LM} & \textbf{1.3B} & \textbf{93.5} & \textbf{94.5} & \textbf{91.6} & \textbf{84.9} & 88.8 & \textbf{79.3} & \textbf{86.7} & \textbf{87.4} \\

    \midrule[0.7pt]

    DS-VL2 & 200B+ & 95.1 & 96.7 & 95.1 & 91.2 & 94.9 & 87.4 & 92.8 & 92.9 \\
    
    \bottomrule[1.1pt]
    \end{tabular}
}
  \vspace{-2mm}

\label{tab:rec_result}
\end{table}

\begin{figure*}[ht]
    \vspace{-9mm}
    \centering
    \includegraphics[width=0.9\textwidth, 
    trim=0in 1in 0in 0.5in, clip
    ]{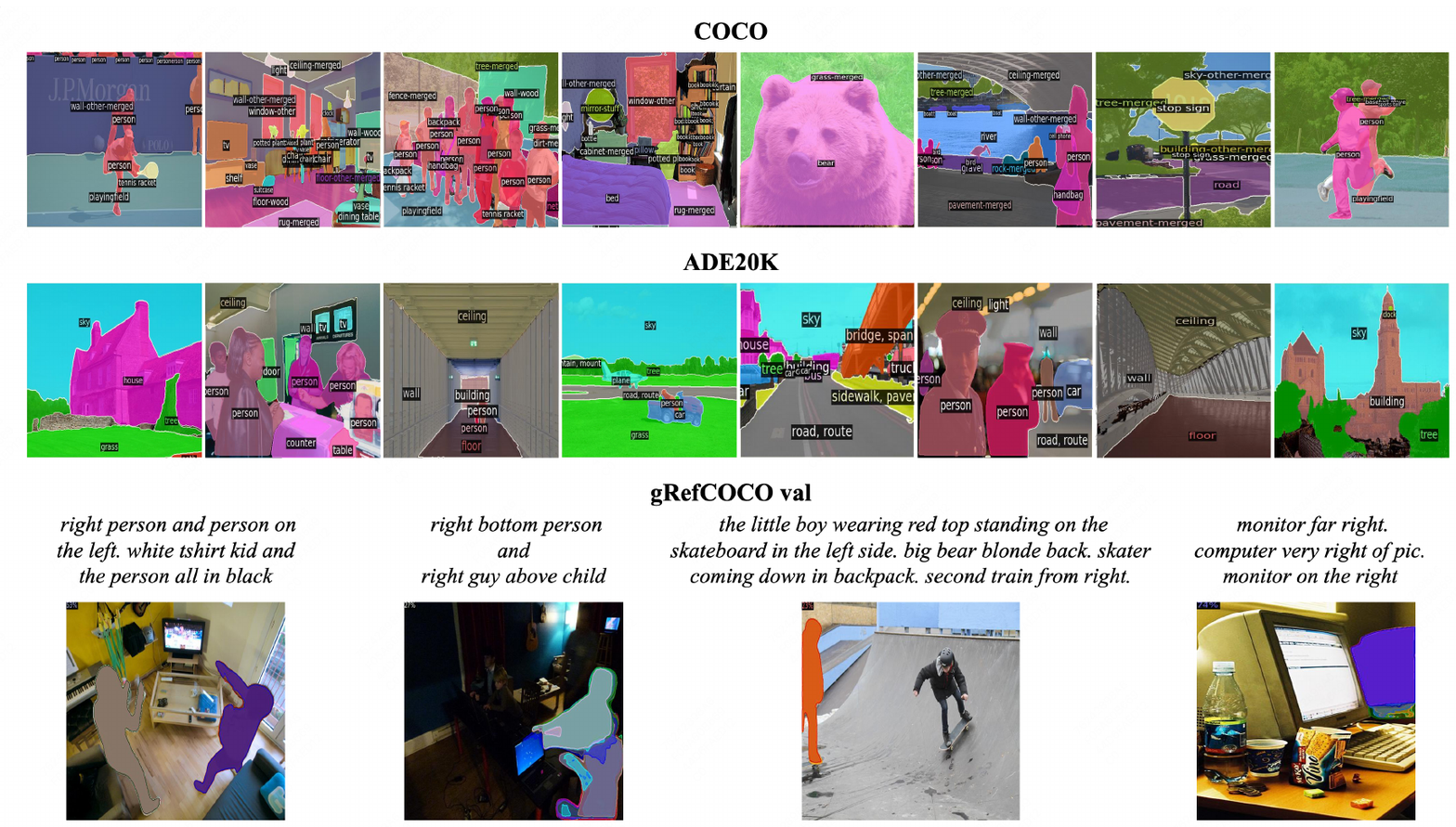}
    \caption{\textbf{Visualization of the segmetation results on COCO, ADE20K and gRefCOCO datasets.} We plot the predicted mask with a confidence larger than 0.5 for a clear visualization. The text above each image in the gRefCOCO denotes the corresponding reference text.}
    \label{fig:seg_visualization}
    \vspace{-3mm}
\end{figure*}

\subsection{Ablation study}
This section presents an ablation study investigating critical architectural decisions, specifically: multi-dataset joint learning, performance across diverse response configurations, the impact of query quantity, and the effect of query selection.

\noindent \textbf{Joint Training.} The progressive incorporation of training data sources is analyzed in \cref{tab:ab-datasets}. Initial training on COCO and RefCOCO datasets yields a \name~mIoU of 77.6 on the RefCOCO validation set. Subsequent inclusion of O365 significantly increases cIoU to 81.8, with further enhancement to 83.6 cIoU upon integrating the Grounding dataset.

For COCO evaluation, sequential integration of O365 yields a PQ of 55.8 and mIoU of 65.9. Subsequent integration of the Grounding dataset marginally improves PQ to 56.1 and mIoU to 66.8.

Crucially, sentence-based perception is enhanced by this multi-dataset approach by 6.0 points while word-based perception remains consistent. These findings empirically validate the effectiveness of our unified multi-task training strategy in coordinating multi-modal representations and significantly enhancing cross-task generalization.

\begin{table}[t]
  \centering
    \setlength{\belowcaptionskip}{-5pt} 
    \setlength{\abovecaptionskip}{5pt} 
  \caption{ Ablation of the datasets involved in the joint training. We use the abbreviations C, R, O, and G to represent COCO, RefCOCO, O365, and Grounding, respectively. Both experiments are trained for 56k iterations.
  }
  \scalebox{0.73}{
    \begin{tabular}{c|c|cc}
    \toprule[1.1pt] 
    \multirow{2}{*}{ Dataset} & 
    \multicolumn{1}{c|}{  RefCOCO val } &
    \multicolumn2{c}{  COCO }  \\
    &cIoU &PQ & mIoU  \\
    \midrule[0.7pt]
    
     C, R & 77.6 & \textbf{56.4} & 66.3\\
     C, R, O & 81.8 & 55.8 & 65.9 \\
     C, R, O, G & \textbf{83.6}  & 56.1 & \textbf{66.8}\\ 
    \bottomrule[1.1pt]
    \end{tabular}
    }
  \vspace{-2mm}
\label{tab:ab-datasets}
\end{table}

\noindent \textbf{Response Settings.} 
We provide a comprehensive ablation study on various response settings in \cref{tab:ab-differenct_cot_styles}. The baseline response, \textit{"The perception result is \textless PER\textgreater"}, remains consistent across all inputs, achieving 75.6 cIoU on the RefCOCO validation set and 55.3 PQ/65.7 mIoU on COCO. When the LLM is instructed to output existing category names or object descriptions from the input image prior to the baseline response, the model maintains its performance on RefCOCO (75.6 cIoU) but exhibits a decline in COCO metrics, resulting in 54.9 PQ and 65.6 mIoU. Conversely, our final configuration, which directs the LLM to generate the image caption before delivering the baseline response, results in simultaneous improvements across all metrics (75.7 cIoU, 55.6 PQ, and 66.2 mIoU).

Additionally, we evaluate the performance of utilizing the most similar visual features as the final queries. Compared to the approach of combining these features with the base query vectors generated by the LLM, this configuration yields a modest improvement in the RefCOCO validation set (+0.2). However, it is accompanied by a noticeable decline in COCO PQ, resulting in a decrease of 0.6.

The incorporation of descriptive captions in the responses facilitates the mutual enhancement of both sentence-level (RefCOCO) and word-level (COCO) perception tasks. This observation verifies our hypothesis that structured semantic prompting significantly improves visual-language understanding.

\begin{table}[t]
  \centering
  \caption{ Ablation of different response settings. 'None' denotes the basic response setting containing the summary token only. 'Existing Obj' means generating existing objects' names or descriptions in the image before summary tokens. 'Caption' indicates the regression of the image caption prior to summary tokens. $^{\dagger}$ denotes the direct use of similar visual features as the final queries without LLM-generated base query vectors.
  }
  \scalebox{0.73}{
    \begin{tabular}{c|c|cc}
    \toprule[1.1pt] 
    \multirow{2}{*}{ Response setting} & 
    \multicolumn{1}{c|}{  RefCOCO val } &
    \multicolumn2{c}{  COCO }  \\
    &cIoU &PQ & mIoU  \\
    \midrule[0.7pt]
    
     None & 75.6 & 55.3 & 65.7\\
     Existing Obj. & 75.6& 54.9 & 65.6 \\
     \textbf{Caption} &  75.7 & \textbf{55.6} & \textbf{66.2} \\ 
     Caption$^{\dagger}$ & \textbf{75.9}& 55.0& 66.1\\
    \bottomrule[1.1pt]
    \end{tabular}
    }
  \vspace{-2mm}
\label{tab:ab-differenct_cot_styles}
\end{table}

\noindent\textbf{Number of Queries.} As presented in \cref{tab:ab-query_number}, the response to variations in the number of queries differs between the RefCOCO and COCO datasets. Specifically, performance on the RefCOCO validation set deteriorates with an increasing number of queries, decreasing from 76.3 cIoU to 72.4 cIoU. We hypothesize that, due to the presence of only one target in the RefCOCO images, an increase in the number of candidates may complicate the model's task. In contrast, for the COCO dataset, the use of 100 queries yields optimal performance among the three settings evaluated. Additionally, it is important to note that an increased number of queries incurs additional computational costs. Consequently, we adopt 100 queries as our default setting to achieve an optimal balance between performance and efficiency.

\noindent\textbf{Query Selection.} As conveyed in \cref{tab:ab-query_selection}, applying separate learnable query embeddings instead of query selection improves RefCOCO val results (from 75.7 to 78.3) but lowers COCO-Panoptic results (from 55.6 to 54.4). It reflects that RefCOCO's simpler targets suit fixed queries, and that COCO's complexity needs more flexible query selection.

\begin{table}[t]
  \centering
  \caption{Ablation of the number of queries. 100 queries is the default setting, considering the balance between performance and computational costs.}
  
  \scalebox{0.73}{
    \begin{tabular}{c|c|cc}
    \toprule[1.1pt] 
    \multirow{2}{*}{ \# queries} & 
    \multicolumn{1}{c|}{  RefCOCO val } &
    \multicolumn2{c}{  COCO }  \\
    &cIoU &PQ & mIoU  \\
    \midrule[0.7pt]
    
     30 & 76.3 & 53.1 & 65.2\\
     \textbf{100} & 75.7 & 55.6 & 66.2  \\
     300 & 72.4 & 54.9 & 65.5\\ 
    \bottomrule[1.1pt]
    \end{tabular}
    }
  \vspace{-2mm}
\label{tab:ab-query_number}
\end{table}

\begin{table}[t]
  \centering
  \caption{ Ablation of whether applying query selection.
  }
  \scalebox{0.73}{
    \begin{tabular}{c|c|cc}
    \toprule[1.1pt] 
    \multirow{2}{*}{ Apply Query Selection } & 
    \multicolumn{1}{c|}{  RefCOCO val } &
    \multicolumn2{c}{  COCO }  \\
    &cIoU &PQ & mIoU  \\
    \midrule[0.7pt]
     \checkmark & 75.7 & 55.6 & 66.2\\
     & 78.3 & 54.4 & 65.1\\
    \bottomrule[1.1pt]
    \end{tabular}
    }
  \vspace{-6mm}
\label{tab:ab-query_selection}
\end{table}

\subsection{Qualitative Results}
We present visualizations for the COCO, ADE20K, and gRefCOCO datasets in \cref{fig:seg_visualization}. The results demonstrate that \name~effectively generates precise masks and is proficient in identifying a comprehensive range of objects.

\section{Conclusion}
In this study, we categorize perception tasks into four groups based on two dimensions: prediction type and instruction type. We observe that current works typically address only a subset of the possible combinations. To address this limitation, we propose a unified Visual Large Language Model framework, \name, encompassing all these tasks. By incorporating a multi-granularity decoder and a CoT-inspired dataset unification strategy, our model enables joint training across multiple datasets and adopts a "thinking-then-perceiving" paradigm. Furthermore, we design a query enhancement strategy to exploit the decoding and generative capabilities of VLLM. Extensive experiments demonstrate the potential across various benchmarks. Considering the improvement brought by our CoT-inspired data curation, we would investigate the application of R1-like training in the realm of perception in our future work.

\section{Acknowledgments}
This work was partly supported by the National Key Research and Development Program of China (No. 2024YFB2808903), the Shenzhen Science and Technology Program (JSGG20220831093004008).

\clearpage

{
    \small
    \bibliographystyle{ieeenat_fullname}
    \bibliography{main}
}

\end{document}